\documentclass[runningheads]{llncs}
\usepackage{eccv}
\usepackage{eccvabbrv}
\usepackage{xspace}
\usepackage{pifont}
\usepackage{graphicx}
\usepackage{booktabs}
\usepackage{adjustbox}
\usepackage[inline]{enumitem}
\newcommand{\cmark}{\ding{51}}
\newcommand{\xmark}{\ding{55}}
\usepackage{multirow,multicol}
\usepackage[accsupp]{axessibility} 
\usepackage[pagebackref,breaklinks,colorlinks,citecolor=eccvblue]{hyperref}
\usepackage[capitalize]{cleveref}
\crefname{section}{Sec.}{Secs.}
\Crefname{section}{Section}{Sections}
\Crefname{table}{Table}{Tables}
\crefname{table}{Tab.}{Tabs.}
\def\benchmark{HOI-Ref\xspace} 
\def\hobenchmark{HO-Ref\xspace} 
\def\hoibenchmark{I-Ref\xspace} 
\def\model{VLM4HOI\xspace} 
\def\dataset{HOI-QA\xspace} 
\definecolor{self_red}{HTML}{FF0000}
\definecolor{royal_blue}{HTML}{4169E1}
\definecolor{green}{HTML}{008000}
\definecolor{orange}{HTML}{FFA500}
\definecolor{pink}{HTML}{FF1493}
\definecolor{gray}{HTML}{C0C0C0}
\definecolor{neon_blue}{HTML}{40E0D0}
\usepackage{orcidlink}
\begin{document}
\title{\benchmark: Hand-Object Interaction Referral in Egocentric Vision}
\titlerunning{HOI-Ref: Hand-Object Interaction Referral}
\author{\href{https://sid2697.github.io/}{\textcolor{black}{Siddhant Bansal}}\and
\href{https://mwray.github.io}{\textcolor{black}{Michael Wray}}\and
\href{https://dimadamen.github.io/}{\textcolor{black}{Dima Damen}}}
\authorrunning{S. Bansal et al.}
\institute{University of Bristol, UK\\
\href{https://sid2697.github.io/hoi-ref/}{https://sid2697.github.io/hoi-ref}}
\maketitle
\begin{abstract}
Large Vision Language Models (VLMs) are now the de facto state-of-the-art for a number of tasks including visual question answering, recognising objects, and spatial referral.
In this work, we propose the \benchmark task for egocentric images that aims to understand interactions between hands and objects using VLMs.
To enable \benchmark, we curate the \dataset dataset that consists of $3.9$M question-answer pairs for training and evaluating VLMs.
\dataset includes questions relating to locating hands, objects, and critically their interactions (\eg referring to the object being manipulated by the hand).
We train the first VLM for \benchmark on this dataset and call it \model.
Our results demonstrate that VLMs trained for referral on third person images fail to recognise and refer hands and objects in egocentric images.
When fine-tuned on our egocentric \dataset dataset, performance improves by $27.9\%$ for referring hands and objects, and by $26.7\%$ for referring interactions.
\end{abstract}

\section{Introduction}
\label{sec:intro}

Understanding hand-object interactions from images~\cite{narasimhaswamy2020detecting,narasimhaswamy2019contextual} and videos~\cite{cheng2023towards,shan2020understanding}, including egocentric videos~\cite{VISOR2022,grauman2022ego4d,ragusa_MECCANO_2023}, offers a fine-grained perception beyond recognising objects in isolation. 
The capability of referring hands as well as objects they are interacting with opens up a range of possibilities for robotics and augmented-reality (AR) applications.
For example, while cooking, current high-end AR glasses are only capable of adding timers over the cookware~\cite{wsj2024visionpro}.
To make these devices truly assistive, we need frameworks that are capable of understanding hand-object interactions from an egocentric perspective.

\begin{figure}[!t]
    \centering
    \includegraphics[width=\textwidth]{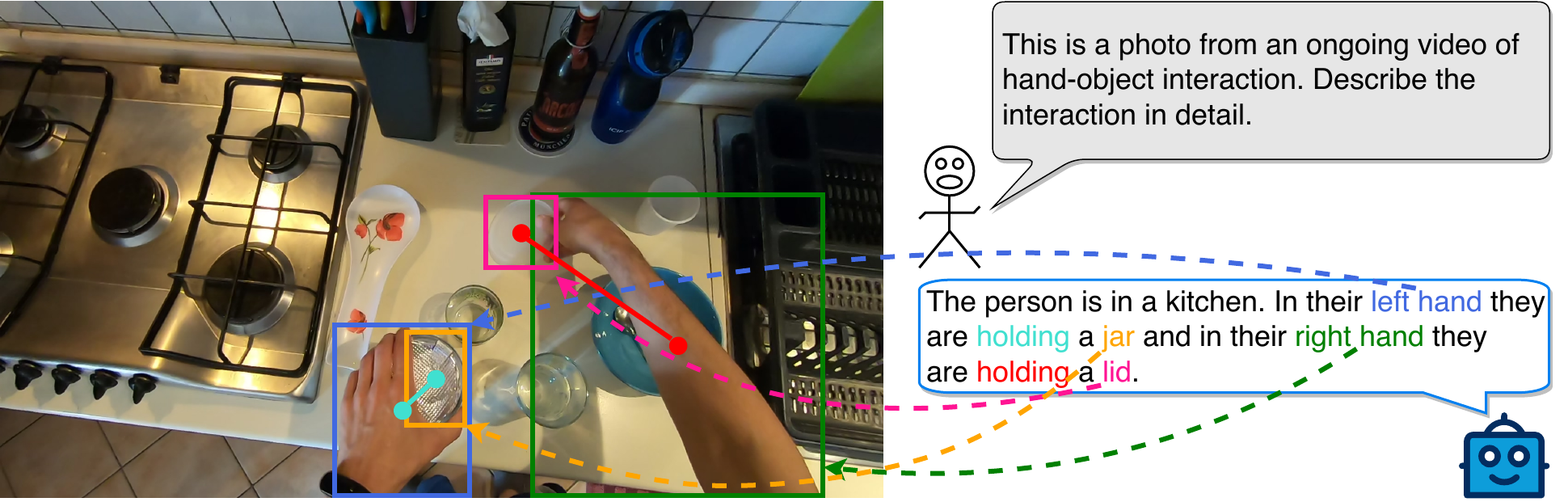}
    \caption{\textbf{Hand-Object Interaction Referral}.
    Given an image from an egocentric video, the goal here is to refer the hands and the objects being interacted with.
    For example, here we wish to refer the \textcolor{royal_blue}{left} and \textcolor{green}{right hand} along with the two objects (\textcolor{orange}{jar} and \textcolor{pink}{lid}) that the hands are interacting with.
    }
    \label{fig:teaser}
\end{figure}

Motivated by the need to assess and improve hand-object interaction understanding, in this work, we introduce \textbf{\benchmark}, \textbf{H}and-\textbf{O}bject \textbf{I}nteraction \textbf{Ref}erral task for egocentric images.
Referral\footnote{The terms referral, grounding, and localisation have been used interchangeably in prior work. In this paper, we consistently use the term referral for this task.} is the task of locating an object in an image---typically using a bounding box, or to name an object given its bounding box~\cite{chen2023shikra,chen2023minigptv2,kosmos_2}.
As shown in \cref{fig:teaser}, given an image from an egocentric video, the goal is to: (a) refer the hands and objects being interacted with and (b) understand the interaction between them.
For example, the \textcolor{royal_blue}{left hand} is \textcolor{neon_blue}{holding} the \textcolor{orange}{jar} and the \textcolor{green}{right hand} is \textcolor{red}{holding} the \textcolor{pink}{lid}.
We solve this by creating a \textit{unified model} capable of understanding spatial coordinates, hands, objects, and hand-object interactions.
To this end, we explore Vision Language Models (VLMs) for hand-object interaction referral in egocentric images.

Thanks to the availability of open-source Large Language Models (LLMs) that are trained on huge amounts of textual data~\cite{chiang2023vicuna,zhang2023llama,zeng2022glm}, training VLMs has recently flourished.
Training regimes (such as fine-tuning a projection layer~\cite{liu2023visual}) have enabled LLMs to generalise well to a range of input queries for solving visual tasks~\cite{awadalla2023openflamingo,dai2023instructblip,liu2023visual,alayrac2022flamingo}.
A few works have looked into referential dialogue~\cite{chen2023shikra} and object referral~\cite{chen2023minigptv2}, primarily using large-scale datasets~\cite{kosmos_2,liu2023visual}.
We build upon these works that utilise LLMs for referring to various objects present in the scene and modify the prompt to perform multiple tasks.
Therefore, we propose our \model model, which enables \benchmark on egocentric images using VLMs.

To train and evaluate \model, we curate a new egocentric dataset (\dataset) of 3.9M question-answer pairs.
 
We construct \dataset by creating a general pipeline that converts available annotations from egocentric datasets to referral question-answer pairs for both training and evaluation.
For \dataset, we utilise the annotations from~\cite{Damen2022RESCALING,grauman2022ego4d,VISOR2022}, where masks, bounding boxes or hand-object interactions have been annotated.
To the best of our knowledge, \dataset is the largest effort to enable hand-object interaction referral for egocentric vision.

Finally, we evaluate \model on two aspects of \benchmark, its ability to refer hands and objects and
 its capability to understand the interaction between the hand and the interacted object.
To summarise, our contributions are:
\begin{enumerate}
    \item We propose the \textbf{\benchmark} task for Egocentric Vision to enable hand-object interaction referral;
    \item We collect the \textbf{\dataset} dataset consisting of $3.9$M question-answer pairs to train and evaluate models for the \benchmark task;
    \item We train \textbf{\model} on \dataset as the first VLM for \benchmark, showcasing its ability to better understand hand-object interactions compared to baselines.
    
\end{enumerate}
Our \dataset dataset, \model model, and code are released on \href{https://sid2697.github.io/hoi-ref/}{project's webpage}  to encourage research on \benchmark task for egocentric vision.
Next, we review recent literature in \cref{sec:related_works}; before presenting details of our model \model and dataset collection details of \dataset in \cref{sec:method}; evaluate and discuss our findings in \cref{sec:experiments}; and provide conclusions and future work in \cref{sec:conclusion}.

\section{Related Works}
\label{sec:related_works}

We first review Large Language Models, before introducing how they have been combined into Vision Tasks.
Next, we present how Vision Language Models have tackled referral before  reviewing their usage for Egocentric Vision.

\paragraph{Large Language Models (LLMs)} have seen a huge surge of popularity~\cite{brown2020language,chowdhery2023palm,raffel2020exploring,zhang2022opt} due to their generative, reasoning, and general knowledge abilities.
These methods have been expanded via instruction-tuning~\cite{ouyang2022training,wang2022benchmarking,wang2022self}, creating instruction-tuned versions of previous LLMs~\cite{achiam2023gpt,chiang2023vicuna,ouyang2022training,zeng2022glm}, allowing for better generalisation for downstream and zero-shot tasks including summarisation, text generation, or question answering.

\paragraph{Vision Language Models (VLMs)} have tackled a variety of downstream tasks such as retrieval, captioning, and visual question answering.
Current VLMs combine LLMs with powerful vision-language representations~\cite{awadalla2023openflamingo,chen2023video,dai2023instructblip,dou2022coarse,li2023videochat,liu2023internchat,shen2023hugginggpt,ye2023mplug,zhu2023minigpt,wu2023visual,yang2023mm,zhang2023llama} via prompt-engineering~\cite{chen2023video}, prompt-tuning~\cite{li2023videochat,liu2023internchat,wu2023visual,yang2023mm}, and instruction tuning~\cite{awadalla2023openflamingo,dai2023instructblip,liu2023visual,ye2023mplug,zhang2023llama,zhu2023minigpt}.

Commonly, models combine a VIT encoder trained from CLIP~\cite{chen2023video,radford2021learning} or BLIP~\cite{li2022blip} with an LLM such as Vicuna~\cite{chiang2023vicuna} or LLAMA~\cite{zhang2023llama}.
These models are later instruction-tuned using multi-modal data to further improve their effectiveness for downstream tasks.
We similarly use instruction tuning to improve the performance of our model.
However, instead of using an LLM to create the instructions~\cite{zhu2023minigpt}, we use rich information such as bounding boxes, dense narrations, and object interactions to generate high quality instructions for the \dataset dataset.

\paragraph{Referral in VLMs}
goes beyond captioning and question answering to referring tasks~\cite{bai2023qwen,chen2023minigptv2,chen2023shikra,liu2023visual,munasinghe2023pg,kosmos_2,guo2024regiongpt}.
These models similarly combine vision encoders with LLMs using LoRA~\cite{hu2022lora} or a projection layer to reduce computation required for training.
Instructions are augmented to include bounding boxes to refer answers within the image or name an object given a bounding box.
Going beyond including tags for bounding boxes in the prompt, MiniGPT-v2~\cite{chen2023minigptv2} proposes using unique identifiers for question types which helps disambiguate the tasks and improves performance on downstream tasks.

We similarly generate instructions in the form of question-answer pairs, focusing on the unique hand-object interactions present within egocentric images. 
We adopt the question identifiers proposed in~\cite{chen2023minigptv2}, to disambiguate task types for the \dataset dataset.
Moreover, we use annotations within egocentric datasets~\cite{Damen2022RESCALING,Damen2018EPICKITCHENS,grauman2022ego4d} to design instructions so that \model can learn to refer not only objects but also interactions in egocentric images.

\paragraph{VLMs for egocentric vision}
have at their onset~\cite{cheng2023can,dai2024gpt4ego,pramanick2023egovlpv2,qinghong2022egocentric,zhao2023learning} focused on video-text retrieval~\cite{Damen2022RESCALING}, natural language queries~\cite{grauman2022ego4d}, and zero-shot action recognition tasks~\cite{Damen2022RESCALING,kuehne2011hmdb,li2018eye,sigurdsson2018charades,soomro2012ucf101}.
These models are trained on the massive-scale Ego4D dataset~\cite{grauman2022ego4d} using crowd-sourced narrations to learn a video-text embedding.
EgoVLP~\cite{qinghong2022egocentric} proposed a multi-choice question benchmark set to verify the performance of the VLM during training.
Follow up works have improved performance via the use of LLMs to narrate the videos as an additional supervision~\cite{zhao2023learning} and a cross-modal fusion layer within the backbone~\cite{pramanick2023egovlpv2}.

More recently, models have been trained for downstream tasks including object tracking via referring expressions~\cite{kurita2023refego} and task verification~\cite{hazra2023egotv}.
RefEgo~\cite{kurita2023refego}, proposes a new dataset for referring object expression which requires models to refer an object from a description across frames of a video along with a new model, solely focused on this task.
Their model is based upon M-DETR~\cite{kamath2021mdetr} modified to cope with objects disappearing from view.
We instead target hand-object interaction referral in frames, going beyond finding objects.
Additionally, our benchmark, and subsequent model, allow for bounding boxes to be used as input as opposed to solely as output.

Up to our knowledge, we are the first to introduce the \benchmark task, propose the \model model to tackle it along with the \dataset dataset for training and evaluation.

\section{Method: Hand-Object Interaction Referral}
\label{sec:method}
Our objective is to enable Vision Language Models (VLMs) to refer hands, objects, and their interaction---the ability to refer the object in contact with the hand, or the hand in contact with the object.
To this end, our work builds on VLMs that take in visual input, in the form of an image along with a text query, and can then generate the appropriate response to refer objects and interactions. 

\begin{figure}[t]
    \centering
    \includegraphics[width=\textwidth]{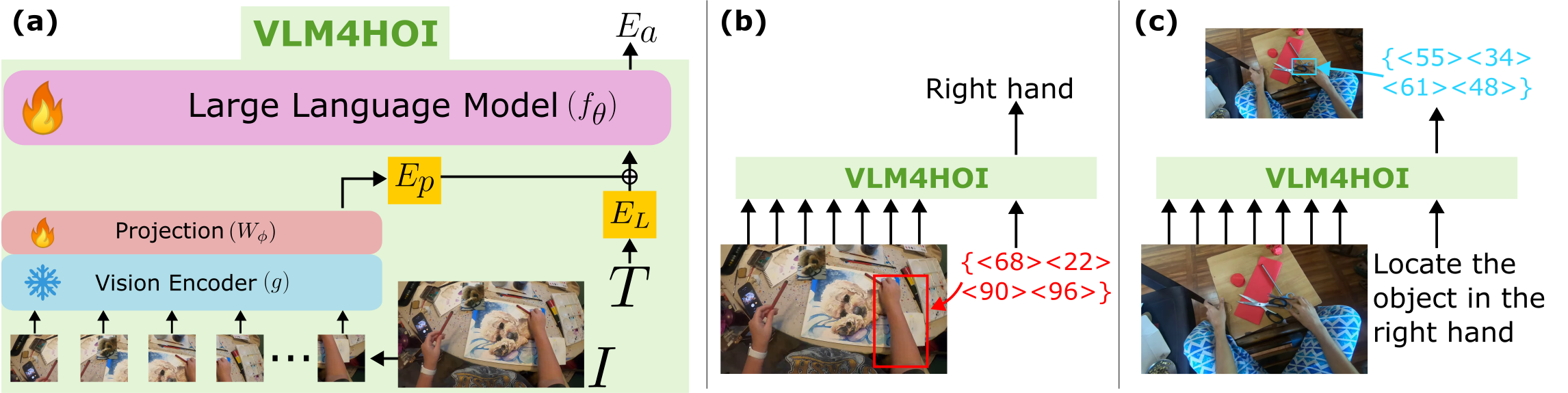}
    \caption{(a) \textbf{\model} for hand-object interaction referral in egocentric images.
    The \model model takes in an image ($I$), passes it through a vision encoder ($g$) and a projection layer ($W_\phi$) to obtain embeddings ($E_p$) in language model's ($f_\theta$) embedding space.
    This is concatenated with the tokenised text ($E_L$) and passed through $f_\theta$ to generate a language response ($E_a$).
    We show two examples where based on the task instruction template, the model generates an output.
    (b), the model identifies a bounding box input as the right hand.
    (c), the model takes in the image and a question to refer the object being held in the right hand and outputs a bounding box.}
    \label{fig:methodology}
\end{figure}

\subsection{\model: VLM for Referral}
\label{subsec:training}

As in prior works~\cite{chen2023minigptv2,chen2023shikra,liu2023visual}, we rely on the abilities of a pre-trained LLM, and integrate the visual tokens into the embedding space of the language model. More specifically, we encode an image $I$ using a visual encoder $g$, a linear projection layer $W_\phi$ and a large language model $f_\theta$.
We explore two types of instructions, one where the instructions include a bounding box to be described, and another where the bounding box is generated in the response.

In detail, we pass the image $I$ through a frozen vision encoder $g$ to obtain patch embeddings $g(I)$.
We use an additional trainable projection layer $W_\phi$ that projects $g(I)$ to a language model's $f_\theta$'s embedding space:
\begin{equation}
    E_p = W_\phi \cdot g(I)
\end{equation}
Once we have the visual tokens ($E_p$) in $f_\theta$'s embedding space, the instruction ($T$) is tokenised ($E_L$) and passed along with $E_p$ through $f_\theta(\cdot)$ to generate a response $E_a$:
\begin{equation}
    E_a = f_\theta(E_p \oplus E_L)
\end{equation}
where $\oplus$ represents concatenation, and $E_a$ is further decoded using the tokeniser to obtain the language response.  

To train the model using language instructions, following~\cite{chen2023shikra,chen2023minigptv2}, we encode the bounding box using the format: 
${\{<X_{\text{left}}><Y_{\text{top}}><X_{\text{right}}><Y_{\text{bottom}}>\}}$, for example, \{<68><22><90><96>\} in \cref{fig:methodology}.
This enables the model to perform referral without utilising a specialised template.
As shown in Fig~\ref{fig:methodology}, we fine-tune the parameters $\theta$ and $\phi$ on a large number of referral examples.

Having presented our model, we now introduce the training data for the tasks of direct referral---that is the referral of hands and/or objects as well as interaction referral---the referral of an object in contact with the hand, and the hand in contact with the object.
Once trained, our model performs referral for hands, objects, and interactions. We thus refer to our model as \model.

\subsection{Framework to Generate Question-Answer Pairs}
\label{subsec:qa_gen}
To enable hand-object referral and understand the interaction between hands and objects using VLMs, question-answer pairs with hand-object referral information are crucial.
In this section, we propose a framework that identifies publicly available annotations that can be converted to question-answer pairs for enabling hand-object interaction referral.

\subsubsection{Annotation Types.}
We use commonly provided annotations in egocentric videos including:
\textbf{\textit{Narrations}} to encode the general activity happening in the egocentric image;
\textbf{\textit{Object Detection GT}} bounding boxes with corresponding semantics, in the form of closed vocabulary (classes) or open vocabulary (captions). We use these detections to enhance the referral capability of VLMs and to understand hand-object interaction;
\textbf{\textit{Hand Detection GT}} bounding boxes with corresponding hand sides (right/left) enable referral and distinction between the interactions of the two hand sides;
\textbf{\textit{Hand-object contact GT}} which highlight whether the hand-side is in contact with an object, and if so which object. We utilise these annotations for interaction referral specifically. For example this enables asking questions similar to, \textit{what is in the left hand of the person?} Answering such a question not only requires the model to correctly refer the hand, but also identify the object in that hand.

Any egocentric dataset consisting of one or more of the annotation types mentioned above can be utilised for generating question-answer pairs to train VLMs for hand-object interaction understanding.
\cref{fig:dataset} shows an example of each annotation.

\subsubsection{Question-Answer Pairs.}
\begin{figure}[!t]
    \centering
    \includegraphics[width=\textwidth]{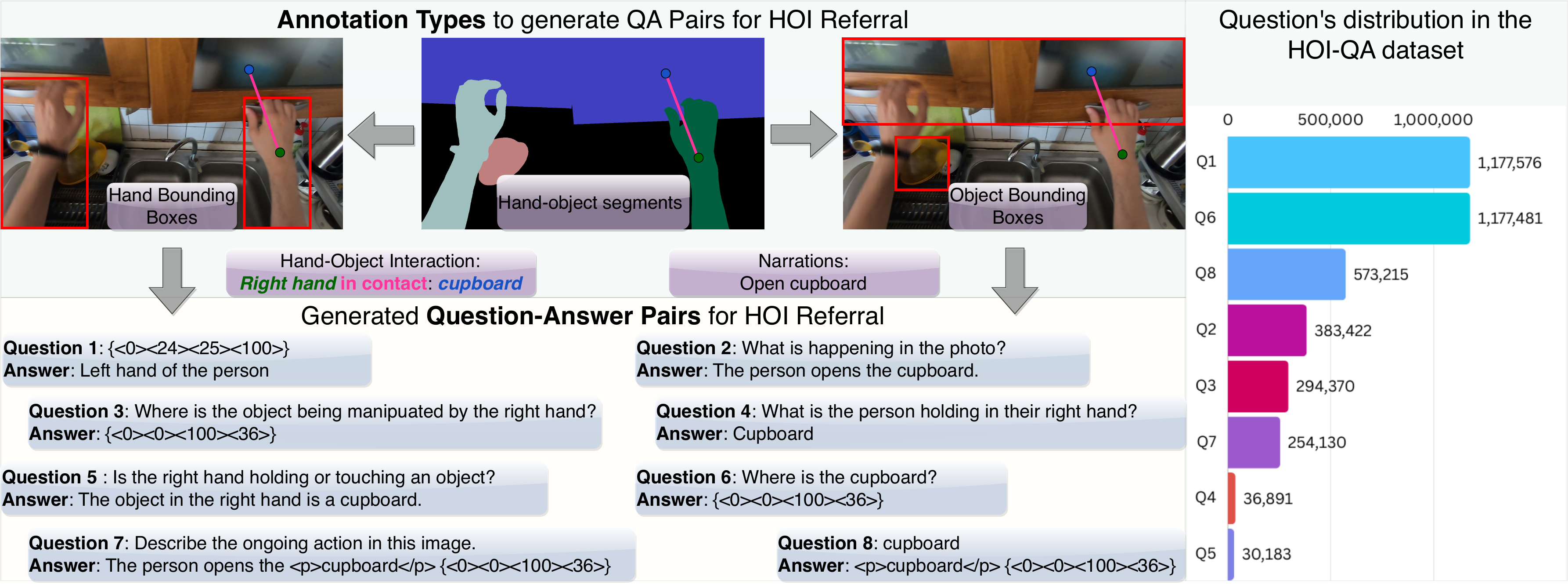}
    \caption{\textbf{Question-Answer pairs} generation for training VLMs to understand hand-object interaction.
    We use multiple annotation types to create the question-answer pairs.
    Top shows the annotations utilised and Bottom shows the types of question-answer pairs generated from these annotations.
    As shown, we convert the segments to bounding boxes to generate various referral questions and utilise contact information to understand interaction between hands and objects.
    Right shows the distribution of questions in the proposed \dataset dataset (\cref{subsec:dataset}).
    }
    \label{fig:dataset}
\end{figure}

Here we discuss how the various annotation types are converted to question-answer pairs.
Note that \textit{\{\textit{object/hand bounding box}\}} in the following discussion corresponds to the bounding box representation shown in \cref{fig:dataset} and discussed in \cref{subsec:training}.
\begin{enumerate}
    \item[\textbullet] \textbf{Narrations} provide us with verb and noun pairs describing the action happening in the egocentric image.
    We convert the verb form to its present tense and formulate the sentence as, ``The person \{verb\} the \{noun\}''.
    This sentence is then used as a description of the event within image.
    For example, \textbf{Question}: \textit{This is a photo from an ongoing video of hand object interactions. What is happening in the photo?} \textbf{Answer}: \textit{The person turns on the tap}. For a visual example, refer to Question $2$ in \cref{fig:dataset}.
    Such QA pairs allow the VLM to have a global understanding of the ongoing action in the image.
    \item[\textbullet] \textbf{Object/Hand Detection Annotations.}
    We use the bounding boxes of hands and objects to create the following QA pairs:
    \begin{enumerate*}[label=(\alph*)]
        \item \textbf{Question}: \textit{Where is the \{object/hand\}?} \textbf{Answer}: \textit{\{Object's/hand's bounding box\}};
        \item \textbf{Question}: \textit{\{Object's/Hand's bounding box\}}. \textbf{Answer}: \textit{\{Object's/Hand's name\}};
        \item \textbf{Question}: \textit{\{Object's name\}}. \textbf{Answer}: \textit{<p>\{Object's name\}</p>} {\{Object's bounding box\}};
        \item \textbf{Question}: \textit{This is a photo from an ongoing video of hand object interactions. Describe the ongoing action in this image.} \textbf{Answer}: \textit{The person opens the <p>tap</p> \{tap's bounding box\}}.
    \end{enumerate*}
    For example, Questions $1$, $6$, $7$, and $8$ in \cref{fig:dataset}.
    Such QA pairs open up the possibility to train VLMs for recognising hands and objects and referring them spatially in the image.
    \item[\textbullet] \textbf{Hand Object Contact information.} We combine bounding box annotations with the knowledge about what object the hand is holding to create the following QA pairs:
    \begin{enumerate*}[label=(\alph*)]
        \item \textbf{Question}: \textit{What is the person holding in their \{left/right\} hand?} \textbf{Answer}: \textit{\{Object's name\}};
        \item \textbf{Question}: \textit{Describe the ongoing action in this image in detail}. \textbf{Answer}: \textit{The person is working in a kitchen. They are holding a <p>\{object's name\}</p> \{object's bounding box\} in their <p>left hand</p> \{left hand's bounding box\} and a <p>\{object's name\}</p> \{object's bounding box\} in their <p>right hand</p> \{right hand's bounding box\}.}
        \item \textbf{Question}: \textit{Is the \{left/right\} hand holding or touching an object?} \textbf{Answer}: \textit{The object in the \{left/right\} hand is a \{object's name\}} OR \textit{\{Left/Right\} hand is not holding anything.}
        \item \textbf{Question}: \textit{Where is the object being manipulated by the hand?} \textbf{Answer}: \textit{\{object's bounding box in left/right hand\}}.
    \end{enumerate*}
    For example, Questions $3$, $4$, and $5$ in \cref{fig:dataset}.
    Such QA pairs not only train VLMs to spatially refer the hands and objects, but also enable them to understand what hand side is manipulating which object.
\end{enumerate}

\subsection{\benchmark: Hand-Object Interaction Referral}
\label{subsec:benchmark}
\begin{figure}[t]
    \centering
    \includegraphics[width=\textwidth]{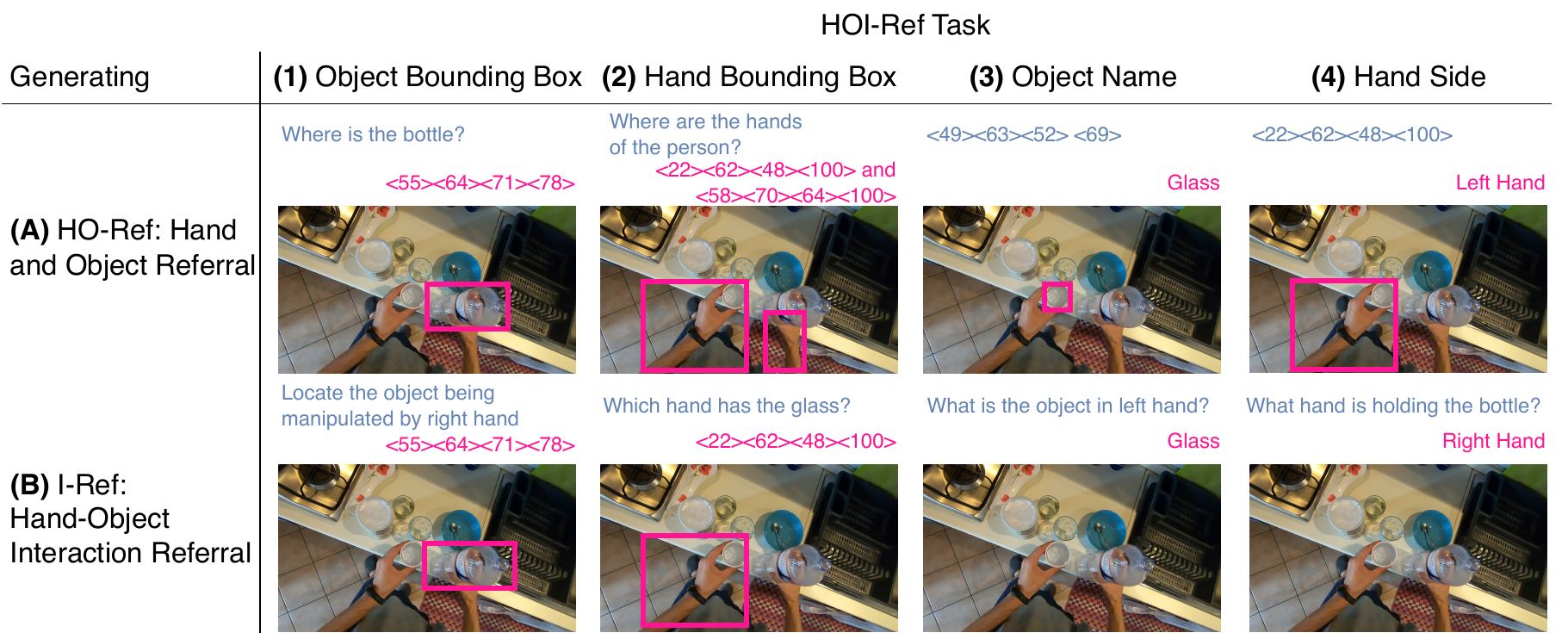}
    \caption{\textbf{\benchmark task} to train and evaluate VLMs for hand-object interaction referral.
    \benchmark focuses on the following two aspects:
    ability to spatially refer and recognise hands and objects and the capability to understand hand-object interaction.
    Columns \textbf{(1)} and \textbf{(2)} evaluate spatially referring hands and objects whereas, columns \textbf{(3)} and \textbf{(4)} aim at object and hand side recognition.
    Moving across rows \textbf{(A)} and \textbf{(B)} shows \benchmark's ability to evaluate for direct referral vs interaction referral.
    For example, in \textbf{A-1}, referring a bottle is simply asking \textit{where is the bottle} however, for \textbf{B-1}, it involves knowing which hand is holding the bottle.
    }
    \label{fig:benchmark}
\end{figure}

In this section, we dive deeper into the formulation of the \textbf{H}and-\textbf{O}bject \textbf{I}nteraction \textbf{Ref}erral (\textbf{\benchmark}) task for egocentric images.
\benchmark aims to create a framework to train and evaluate VLMs on two aspects of hand-object interaction referral:
\begin{enumerate*}[label=(\alph*)]
    \item ability to spatially refer and recognise hands and objects and
    \item capability to understand the interaction between hand and object.
\end{enumerate*}
To achieve this, the question-answer pairs created in \cref{subsec:qa_gen} are strategically divided into two parts (\cref{fig:benchmark}):
\begin{enumerate*}[label=(\Alph*)]
    \item \textbf{\hobenchmark}: \textbf{H}and and \textbf{O}bject \textbf{Ref}erral, and
    \item \textbf{\hoibenchmark}: Hand-Object \textbf{I}nteraction \textbf{Ref}erral.
\end{enumerate*}

\subsubsection{\hobenchmark: Hand and Object Referral.}
The questions in \hobenchmark are further divided into four general categories depending on the input $\rightarrow$ output types:
\begin{enumerate}
    \item \textbf{Object's Name $\rightarrow$ Object's bounding box}. Given the object's name, the goal is to predict its spatial coordinates. The question in this category is: \textit{Where is the \{object's name\}?}
    For example, question \textbf{A-1} in \cref{fig:benchmark}.
    \item \textbf{Hand side $\rightarrow$ Hand's bounding box}. Given the hand's side, the goal is to predict its spatial coordinates. The questions in this category are:
        \textit{Where is the \{left/right\} hand of the person?} Or
        \textit{Where are the hands of the person?}
    For example, question \textbf{A-2} in \cref{fig:benchmark}.
    \item \textbf{Object's bounding box $\rightarrow$ Object's Name}. Given the object's bounding box, the goal is to predict its name. The question in this category is:
    \textit{\{object's bounding box\}}.
    For example, question \textbf{A-3} in \cref{fig:benchmark}.
    \item \textbf{Hand's bounding box $\rightarrow$ Hand side}. Given the hand's bounding box, the goal is to predict the hand's side. The question in this category is:
    \textit{\{left/right  hand's bounding box\}}.
    For example, question \textbf{A-4} in \cref{fig:benchmark}.
\end{enumerate}

\subsubsection{\hoibenchmark: Hand-Object Interaction Referral.}
Similar to above, we have four corresponding categories. However, answering questions for \hoibenchmark requires not only identifying the object/hand but the interaction between this and another hand/object, hence the term Interaction Referral. The four categories are:

\begin{enumerate}
    \item \textbf{Hand's name $\rightarrow$ Object's bounding box}. Given the hand side's name, the goal is to predict spatial coordinates of the object in that hand. The question in this category is:
    \textit{Locate the object being manipulated by \{right/left\} hand.}
    For example, question \textbf{B-1} in \cref{fig:benchmark}.
    \item \textbf{Object's Name $\rightarrow$ Hand's bounding box}. Given an object's name, the goal is to predict spatial coordinates of the hand holding the object. The question in this category is:
    \textit{Which hand has the \{object's name\}?}
    For example, question \textbf{B-2} in \cref{fig:benchmark}.    
    \item \textbf{Hand's name $\rightarrow$ Object's name}. Opposing the previous category, given a hand side, the goal is to generate the object's name in that hand. The question in this category is:
     \textit{What is the object in \{left/right\} hand?}
    For example, question \textbf{B-3} in \cref{fig:benchmark}.
    \item \textbf{Object's name $\rightarrow$ Hand's name}. Given an object's name, the goal is to predict hand side that is holding the object. The questions in this category are:
     \textit{What hand is holding the \{object's name\}?} Or
     \textit{What \{object\} is in the hands of the person?}
    For example, question \textbf{B-4} in \cref{fig:benchmark}.
\end{enumerate}

\noindent \textbf{Task Tags}
\label{subsec:tasktags}
Until now the question-answer pairs generated do not have a marker on them that guides the VLM on what to expect in terms of the style of input/output. 
For example, it might be challenging for a VLM to know what to do with Question \textbf{A-3} in \cref{fig:benchmark}.
To alleviate this issue, following the lead of~\cite{chen2023minigptv2}, we adopt the multi-task instruction template by adding various \textit{task tags} to the questions generated from the process described earlier.
Similar to~\cite{chen2023minigptv2}, we add the task tag at the beginning of the question.
We use \textit{[refer]} for questions where hand or object is referred, these are \textbf{A-1, A-2, B-1} and \textbf{B-2} in \cref{fig:benchmark}.
We use \textit{[identify]} for questions where task is to generate referred hand or object's name.
These are \textbf{A-3} and \textbf{A-4} in \cref{fig:benchmark}.
We use \textit{[vqa]} for questions without bounding boxes, these are \textbf{B-3} and \textbf{B-4} in \cref{fig:benchmark}.

\subsection{\dataset: Dataset for Hand-Object Interaction Referral}
\label{subsec:dataset}
Having established the framework to generate question-answer pairs in \cref{subsec:qa_gen}, and the targeted \hobenchmark and \hoibenchmark from \cref{subsec:benchmark}, we now detail our collection of QA pairs using publicly available egocentric datasets.

Out of various available datasets for egocentric videos~\cite{Damen2022RESCALING,grauman2022ego4d,EgoProceLECCV2022,ragusa_MECCANO_2023,li2018eye,Damen2018EPICKITCHENS,sener2022assembly101}, 
we select EPIC-Kitchens~\cite{Damen2022RESCALING,Damen2018EPICKITCHENS,VISOR2022} and Ego4D~\cite{grauman2022ego4d} as our data source for generating the QA pairs.
These are the largest datasets in egocentric vision and provide crucial annotations including narrations and hand-object interaction information.
This is in contrast with other datasets that do not have bounding box annotations~\cite{charadesego,EgoProceLECCV2022} or hand-object interaction information~\cite{charadesego,EgoProceLECCV2022,li2018eye}.
Other datasets that have the hand-object information are noticeably smaller~\cite{ragusa_MECCANO_2023,sener2022assembly101}. 

EPIC-Kitchens contains all the annotation types mentioned in \cref{subsec:qa_gen}, \ie, narrations~\cite{Damen2018EPICKITCHENS,Damen2022RESCALING}, object bounding boxes~\cite{Damen2018EPICKITCHENS}, and hand-object contact information~\cite{VISOR2022}.
Ego4D contains a large number of narrations and hand and object bounding box annotation types, particularly within the forecasting hands and objects (FHO) benchmark which we select as a subset.

\begin{table}[!t]
    \centering
    \caption{\textbf{Comparison of referral-based question-answering datasets.} Our \dataset dataset has multiple orders of magnitude more questions than the current largest third-person dataset
    }
    \setlength{\tabcolsep}{5pt}
    \begin{adjustbox}{width=1\textwidth}
    \begin{tabular}{@{}lcccc@{}}\toprule
    Dataset & \# QA Pairs & Automatic Generation? & Task Tags? & Egocentric?\\
    \midrule
    Shikra-RD~\cite{chen2023shikra} & $5,922$ & \textcolor{red}{\xmark} & \textcolor{red}{\xmark} & \textcolor{red}{\xmark} \\ 
    Flicker30k~\cite{chen2023minigptv2,7410660} & $2.5$K & \textcolor{green}{\cmark} & \textcolor{red}{\xmark}  & \textcolor{red}{\xmark} \\
    Multi-task Conv.~\cite{chen2023minigptv2} & $60.9$K & \textcolor{green}{\cmark} & \textcolor{green}{\cmark}  & \textcolor{red}{\xmark} \\
    \dataset (ours) & $\mathbf{3.9}$\textbf{M} & \textcolor{green}{\cmark}  & \textcolor{green}{\cmark} & \textcolor{green}{\cmark}\\
    \bottomrule
    \end{tabular}
    \end{adjustbox}
    \label{tab:dataset_stats}
\end{table}

We create \textbf{3.9M question-answer pairs} for the \dataset dataset.
Out of 3.9M QA pairs, 1.8M pairs are from EPIC-Kitchen's and 2.1M pairs are from Ego4D's FHO benchmark.
Based on the train/val splits of EPIC-Kitchens and Ego4D, 
\dataset is divided into $3.2$M QA pairs for train and $748$K for test. 
Out of $748$K QA pairs in the test set, approximately $20\%$ are from EPIC-Kitchens and $80\%$ are from Ego4D.
Note that the split is on the video level, i.e. all QA pairs from the same video are in the same split (train/test).

The scale of the proposed \dataset is suitable for training large models.
In \cref{tab:dataset_stats}, we compare \dataset with existing referential-based question-answering datasets. \dataset contains the highest number of question-answer pairs; is the only dataset for egocentric images; and does not require prompting an LLM~\cite{chen2023shikra} for its creation, resulting in a cleaner set of annotations.
\Cref{fig:dataset} (right) shows the distribution of questions within \dataset.

\section{Experiments}
\label{sec:experiments}

In this section, we first provide details on implementation (\cref{subsec:implementation}); evaluation protocol (\cref{subsec:evaluation}); and baselines (\cref{subsec:baselines}) followed by results of \model on \dataset (\cref{subsec:main_results}), ablations (\cref{subsec:ablation}), and qualitative results (\cref{subsec:qualitative}).

\subsection{Implementation Details}
\label{subsec:implementation}
We use EVA~\cite{10203681} as our vision encoder for \model and keep it frozen during training.
We train the linear projection layer and fine-tune the large language model.
We use LLaMA2-chat (7B)~\cite{Touvron2023Llama2O} as our LLM which is finetuned using LoRA~\cite{hu2022lora} with rank $r=64$.
Following~\cite{chen2023minigptv2}, we train \model with an image resolution of $448\times448$ and before projecting to LLaMA-2's representation space, four adjacent visual output tokens from EVA are concatenated.
We initialise the language model using the weights from the third stage of MiniGPT-v2~\cite{chen2023minigptv2}.
This allows us to finetune a model that has knowledge about referral in third-person images and task tags.
Additionally, we adapt the instruction template from LLaMA-2 as our input to the model:
\textit{[INST] <Img><ImageFeature></Img> [task tag] Instruction [/INST]}.
Here, \textit{ImageFeature} is replaced by the visual features in LLaMA-2's embedding space and the relevant \textit{task tag} and \textit{instruction} are filled in.
Finally, all the bounding boxes in the question-answer pairs are normalised to $(100, 100)$.
We train \model using $8$ Nvidia V100 GPUs for 1 epoch on \dataset that takes approximately $28$ hours.

\subsection{Metrics}
\label{subsec:evaluation}
Based on answer type, we have two evaluation schemes.
For questions involving a noun as answer, we perform word-level matching and calculate accuracy.
For questions generating bounding boxes, we consider the predicted bounding box as correct if its IoU with the ground truth is above $0.5$. 
For questions with multiple bounding boxes in their answers, we consider the answer to correctly only when each ground-truth bounding box has prediction with IoU above $0.5$.

\subsection{Baselines}
\label{subsec:baselines}
To evaluate \model on \dataset we consider the following baselines:
\begin{enumerate*}
    \item \textit{Random}. For noun generation, we select a noun from the distribution of nouns within the training set.
    For bounding boxes, we use the bounding box obtained by calculating the mean value of top-left and bottom-right corners of the bounding boxes over the training set as our prediction.
    \item \textit{MiniGPT-v2~\cite{chen2023minigptv2}}. Trained for referral on third-person images and utilises task tags to flag the type of response expected. This is the closest baseline to \model. We use the publicly available model from the final stage of MiniGPT-v2 as our baseline.
\end{enumerate*}
There are additional works performing referral on third-person images~(\eg~\cite{chen2023shikra,kosmos_2}), however, they do not utilise task tags for a fair comparison and are outperformed on existing datasets by MiniGPT-v2 based on the results reported in~\cite{chen2023minigptv2}.

\subsection{Quantitative Evaluation}
\label{subsec:main_results}
In this section, we first present results of \model on \dataset, before analysing how well \model performs on each task: \hobenchmark and \hoibenchmark.

\vspace{-10pt}
\subsubsection{Results of \model on \dataset}

\Cref{tab:main_results} (left) contains results on the complete test set of \dataset.
The random baseline performs poorly on both noun and bounding box accuracy.
This highlights the challenging nature of \dataset.
Secondly, \model outperforms both the baselines by a large margin.
Specifically, \model achieves $11.08\%$ higher noun accuracy and $27.85\%$ higher bounding box accuracy as compared to MiniGPT-v2~\cite{chen2023minigptv2}.
This shows that VLMs trained on third-person images fall short to generalise on egocentric images, and that improving their performance on egocentric data requires a specialised dataset.
Furthermore, as questions in \dataset require understanding hand-object interactions, the significant difference in performance between MiniGPT-v2 and \model indicates that even large scale third-person training does not enable hand-object interaction understanding and localisation.

By analysing the two subsets of our benchmark (\cref{tab:main_results}, right), \textbf{HO-Ref} vs \textbf{I-Ref}, we note that our model's focus on interaction referral results in a significant increase in performance for noun accuracy during referral ($30.39\%$ on I-Ref compared to $10.10\%$ on HO-Ref). Improvements in bounding box accuracy are comparable.
We provide a detailed breakdown of results on the eight sub-divisions of \hobenchmark and \hoibenchmark in the supplementary.

\begin{table}[!t]
    \centering
    \caption{\textbf{Results on \dataset and analysis on \hobenchmark and \hoibenchmark}.
    Left shows the results on all of \dataset's test set.
    Right compares the performance on \hobenchmark and \hoibenchmark.
    \model outperforms the baselines highlighting its capability to understand hands/objects and refer interactions in an egocentric image.
    Here, N-A stands for Noun Accuracy, BB-A stands for Bounding Box Accuracy, and Avg. IOU stands for Average bounding box accuracy
    }
    \setlength{\tabcolsep}{5pt}
    \begin{adjustbox}{width=1\textwidth}
    \begin{tabular}{@{}lccccccccc@{}}\toprule
    \multirow{2}{*}{Model} & \multicolumn{3}{c}{\dataset} & \multicolumn{3}{c}{\hobenchmark} & \multicolumn{3}{c}{\hoibenchmark}\\
    \cmidrule(lr){2-4}\cmidrule(lr){5-7}\cmidrule(l){8-10}
    & N-A & BB-A & Avg. IOU & N-A & BB-A & Avg. IOU & N-A & BB-A & Avg. IOU \\
    \midrule
    Random & $0.50$ & $1.22$ & $7.00$ & $0.05$ & $1.15$ & $6.77$ & $0.02$ &  $0.07$ & $12.45$ \\
    MiniGPT-v2~\cite{chen2023minigptv2} & $6.70$ & $22.46$ & $24.49$ & $6.68$ & $22.13$ & $24.15$ & $7.06$ & $30.12$ & $32.53$ \\
    \model (ours) & $\mathbf{17.78}$ & $\mathbf{50.26}$ & $\mathbf{43.21}$ & $\mathbf{16.78}$ & $\mathbf{49.98}$ & $\mathbf{42.83}$ & $\mathbf{37.45}$ & $\mathbf{56.86}$ & $\mathbf{52.40}$\\
    \bottomrule
    \end{tabular}
    \end{adjustbox}
    \label{tab:main_results}
\end{table}

\subsection{Qualitative Results}
\label{subsec:qualitative}
\begin{figure}[t]
    \centering
    \includegraphics[width=\textwidth]{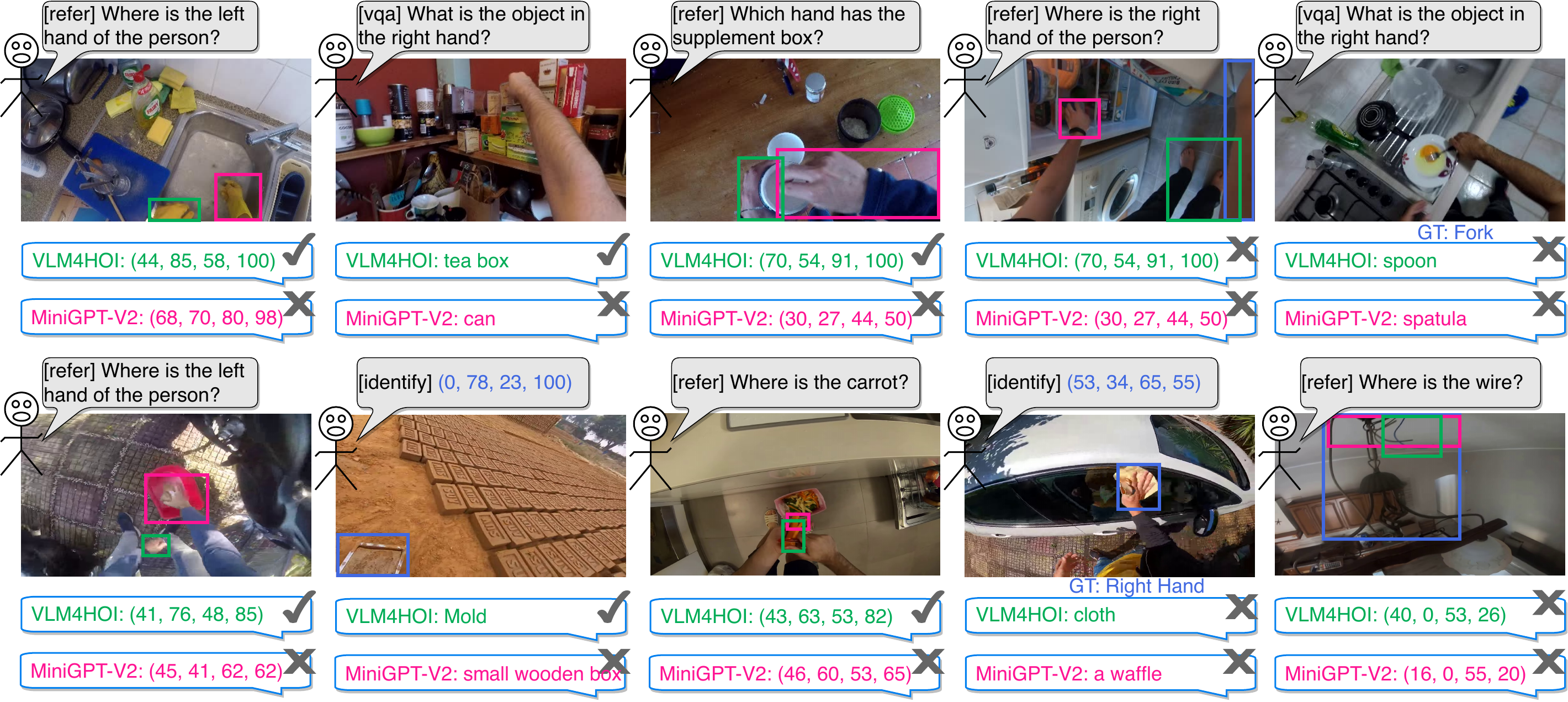}
    \caption{\textbf{Qualitative Results} on \textcolor{green}{\model} and \textcolor{pink}{MiniGPT-v2}~\cite{chen2023minigptv2} on \dataset.
    For questions with correct bounding box output, the ground truth bounding box is omitted.
    When both models are incorrect, we add the ground truth in \textcolor{blue}{blue}.
    \model performs well on most of the cases where MiniGPT-v2 falls short.
    \model fails in case of ambiguity.
    For example, it identifies the hand as cloth as the hand is holding the cloth (MiniGPT-v2 predicts it as a waffle).
    }
    \label{fig:qualitative}
\end{figure}

\Cref{fig:qualitative} shows qualitative results for \model and MiniGPT-v2.
\model performs adequately for the cases shown where MiniGPT-v2 falls short.
\model is capable of referring hands even when in glove (row-1, col-1), identify the exact object in hand amongst others on the shelf (row-1, col-2) and better predict the bounding box for in-hand objects (row-2, col-3). 

\model fails in the ambiguous cases.
For example, when asked about what is in the right hand, \model identifies the fork as a spoon (row-1, col-5) due to the small size of the object as well as motion blur.
When the hand side is missing, both models are confused by the present hand or even feet in view (row-1, col-4). 
At times, the annotations might be confusing themselves.
For example when asked about the bounding box (53, 54, 65, 55) (row-2, col-4), the hand fully capturing the cloth makes a perfectly overlapping bounding box.
\model identifies the object as opposed to the hand.

\vspace{-5pt}
\subsection{Ablation Study}
\label{subsec:ablation}
In this section, we first ablate whether spatial information is important by removing bounding box supervision during training. Next, we showcase the importance of dataset selection of both EPIC and Ego4D data within \dataset. Finally, we evaluate the importance of the unique task tags in training \model.

\begin{table}[!t]
    \centering
    \caption{\textbf{Importance of Spatial Information}.
    Train \model without any spatial supervision. We show that for enabling hand-object interaction referral, spatial information is crucial
    }
    \setlength{\tabcolsep}{5pt}
    \begin{adjustbox}{width=1\textwidth}
    \begin{tabular}{@{}lccc@{}}\toprule
    Model & Noun Accuracy & Bounding Box Accuracy & Avg. Bounding Box IOU \\ 
    \midrule
    w/\text{o} Bbox. & $10.72$ & $8.98$ & $14.74$ \\
    w/\ Bbox. & $\mathbf{17.78}$ & $\mathbf{50.26}$ & $\mathbf{43.21}$ \\
    \bottomrule
    \end{tabular}
    \end{adjustbox}
    \label{tab:nobb_results}
\end{table}

\vspace{5pt}
\subsubsection{Importance of spatial information.}
Here, we study the impact of bounding box supervision in training \model.
We select the subset of question-answer pairs from \dataset that \textit{do not} contain any bounding box supervision.
This leaves us with a subset of $691,380$ QA pairs from Ego4D and $279,458$ QA pairs from EPIC-Kitchens (approximately $971K$ QA pairs) for training \model.
As shown in \cref{tab:nobb_results}, the bounding box accuracy significantly drops across all results.
It is intuitive that without training for spatial localisation, \model will not be able to perform well.
The results however show that the model is unable to leverage its previous training on third-person data and LLaMA-2's large training corpus to locate objects in egocentric images.
Furthermore, the drop in noun accuracy highlights the importance of benefit of localisation supervision to understanding an object's appearance.

\subsubsection{Dataset Contributions.}

As we train on annotations from two datasets, we analyse the contribution of each dataset to the results by training on QA pairs solely from each dataset.
Accordingly, these are: the $1.7$M pairs of \dataset from EPIC-Kitchens; and the $1.5$M QA pairs of \dataset from Ego4D.
As shown in \cref{tab:epictrain_results} Bounding box accuracy drops when we train on only one dataset highlighting that both the datasets play a crucial role for hand-object interaction referral.
The model trained solely on Ego4D performs better as compared to training on EPIC-Kitchens, this is because of the larger and diverse nature of the Ego4D dataset.
However, the noun accuracy drops when we utilise both the datasets. we believe this is because of the bias in our test set being more aligned with the smaller subset of noun classes available in EPIC-Kitchens.

\begin{table}[!t]
    \centering
    \caption{\textbf{Dataset Contributions}.
    Here we train the model on a specific dataset and test on the test set of \dataset.
    This highlights dataset's contribution to the performance.
    }
    \setlength{\tabcolsep}{5pt}
    \begin{adjustbox}{width=1\textwidth}
    \begin{tabular}{@{}lccc@{}}\toprule
    Model & Noun Accuracy & Bounding Box Accuracy & Avg. Bounding Box IOU \\ 
    \midrule
    Only EPIC  & $\mathbf{20.63}$ & $25.45$ & $28.16$\\ 
    Only Ego4D  & $13.71$ & $43.42$ & $38.12$ \\ 
    Full Training  & $17.78$ & $\mathbf{50.26}$ & $\mathbf{43.21}$ \\
    \bottomrule
    \end{tabular}
    \end{adjustbox}
    \label{tab:epictrain_results}
\end{table}

\subsubsection{Importance of Task Tags.}
In this section, we evaluate the importance of the unique task identifiers in the form of task tags in which we train \model with/without the task tags. The results can be seen in \cref{tab:task_tags} where the model with task tags outperforms by a small margin of $0.31\%$ on bounding box accuracy and $0.56\%$ on noun accuracy.
As in~\cite{chen2023minigptv2}, the inclusion of task tags marginally helps the model disambiguate the questions and improves overall performance.

\begin{table}[!t]
    \centering
    \caption{\textbf{Importance of Task Tags}. Here we train and evaluate \model without the task tags. Similar to~\cite{chen2023minigptv2}, we see a slight improvement in the performance
    }
    \setlength{\tabcolsep}{5pt}
    \begin{adjustbox}{width=1\textwidth}
    \begin{tabular}{@{}lccc@{}}\toprule
    Model & Noun Accuracy & Bounding Box Accuracy & Avg. Bounding Box IOU \\
    \midrule
    w/\text{o} Task Tags & $17.22$ & $49.95$ & $43.07$ \\
    w/\ Task Tags & $\mathbf{17.78}$ & $\mathbf{50.26}$ & $\mathbf{43.21}$ \\
    \bottomrule
    \end{tabular}
    \end{adjustbox}
    \label{tab:task_tags}
\end{table}

\vspace{-9pt}
\subsection{Limitations}
\label{subsec:limitations}

There are two sources of bias in \model---the first is inherit bias in the pre-trained LLM. Despite fine-tuning, language biases are known to exist in all VLMs.
Secondly, bias in the \dataset, which corresponds to the natural occurrences of certain actions, objects and interactions over others. 
As \model is trained on unconstrained, long-tailed data of objects and actions~\cite{grauman2022ego4d,Damen2022RESCALING}, it will more commonly predict frequent objects and interactions over infrequent ones.

\model is also reliant on the strength of the visual encoder. \ie Small objects and motion blur can cause errors in the model's prediction and would require a stronger visual encoder or multiple frames from a video.

\vspace{-9pt}
\section{Conclusion}
\label{sec:conclusion}

In this work, we propose and explore the \textbf{\benchmark} task for egocentric vision to enable hand-object interaction referral using VLMs.
We further investigate two aspects of \benchmark, referral of hands and objects (\hobenchmark) and understanding of the interactions between them (\hoibenchmark).
We propose the \dataset dataset that consists of 3.9M question-answer pairs for training and evaluation of VLMs for \benchmark.
We train a \textit{unified} VLM, \model, on \dataset and significantly improve hand-object interaction referral performance over baselines.
The dataset, models, and task proposed will pave the way for hand-object interaction referral in egocentric images.

\subsection*{Acknowledgements}
This work uses public datasets---code and models are publicly available. The research is supported by EPSRC UMPIRE EP/T004991/1 and EPSRC Programme Grant VisualAI EP/T028572/1. S Bansal is supported by a Charitable Donation to the University of Bristol from Meta. We acknowledge the use of the EPSRC funded Tier 2 facility JADE-II EP/T022205/1.

The authors would like to thank Alexandros Stergiou, Kranti Kumar Parida, Samuel Pollard, and Rhodri Guerrier for their comments on the manuscript.

\bibliographystyle{splncs04}
\bibliography{main}

\clearpage
\section{Appendix}
\subsection{Overview}
\label{supp_sec:intro}
In the supplementary, we provide in-depth details of the proposed \dataset dataset (\cref{supp_sec:dataset_details}).
We then provide additional quantitative results in \cref{supp_sec:quantiative_results}, showcasing example questions from both EPIC-Kitchens~\cite{Damen2018EPICKITCHENS} and Ego4D~\cite{grauman2022ego4d} datasets.
Finally, we present zero-shot interactive results using the proposed \model model on the \dataset dataset in \cref{supp_sec:qualitative_results}.

\section{Dataset details}
\label{supp_sec:dataset_details}

\subsection{Additional dataset statistics}

\subsubsection{\dataset distribution details.} 
\begin{figure}[!tbhp]
    \centering
    \includegraphics[width=0.93\textwidth]{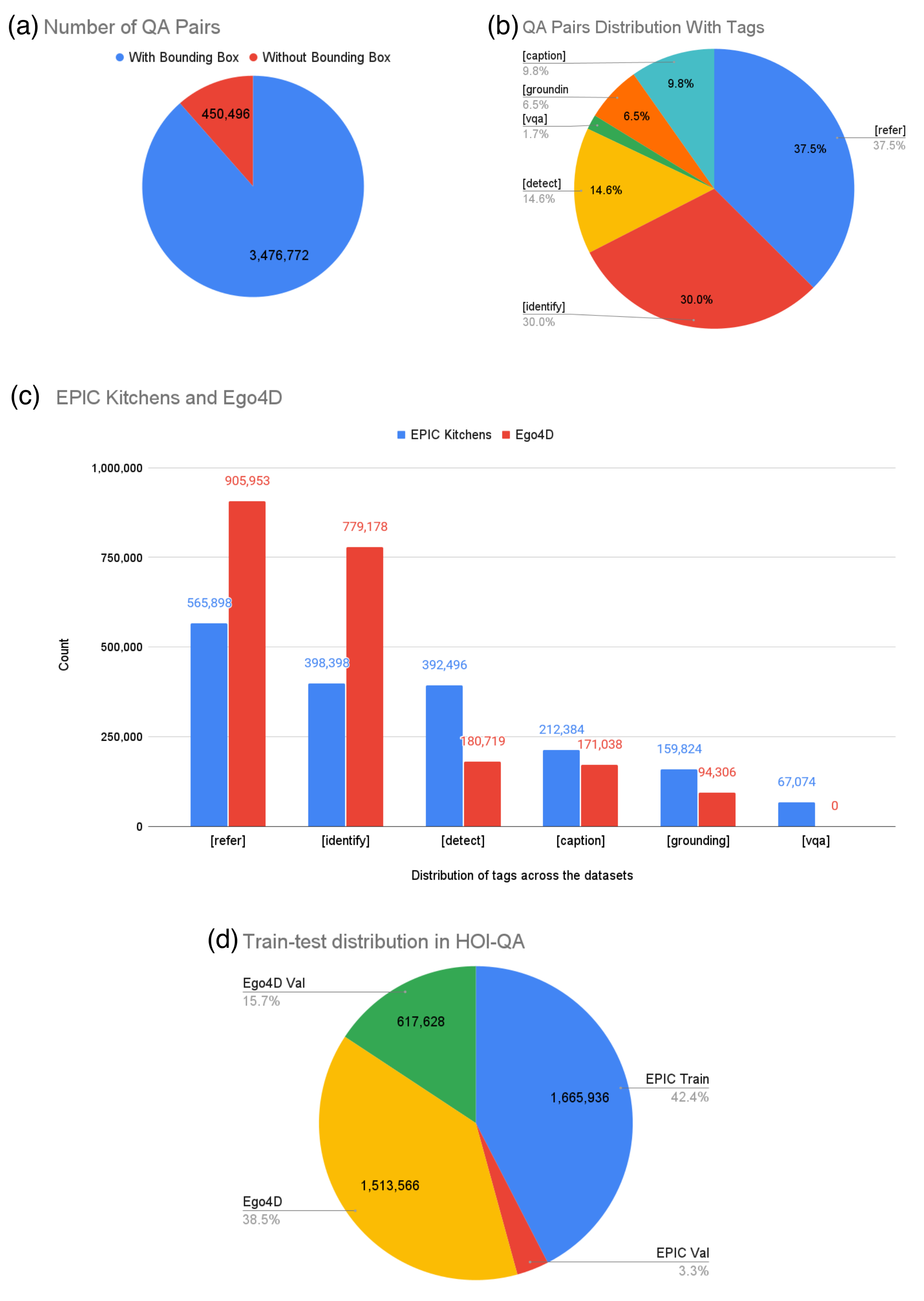}
    \caption{
    \textbf{\dataset distribution details:}
    \textbf{(a)} shows the distribution of QA pairs across the \dataset dataset with and without bounding boxes;
    \textbf{(b)} contains the distribution of QA pairs in \dataset based on the task tags;
    \textbf{(c)} provides the distribution of QA pairs across EPIC-Kitchens and Ego4D; and 
    \textbf{(d)} showcases the distribution of EPIC-Kitchens and Ego4D across the train and test split of \dataset.
    As can be seen, the majority of QA pairs contains bounding boxes to enable hand-object interaction referral.
    }
    \label{supp_fig:distribution}
\end{figure}

In this section, we dive deep into the details of the proposed \dataset dataset.
As mentioned in the main paper, \dataset consists of $3.9$M question-answer pairs to train and test various Vision Language Models (VLMs).
\Cref{supp_fig:distribution} shows different statistics of \dataset regarding the QA pairs.

\Cref{supp_fig:distribution} \textbf{(a)} contains the distribution of QA pairs with and without bounding boxes.
As we particularly target referral, the dataset has a larger number of questions with bounding boxes to enable hand-object interaction referral for VLMs.
In Figure 3 (main paper) we show the distribution of QA pairs based on the question type, here (\cref{supp_fig:distribution} \textbf{(b)}) we additionally show the distribution of task tags within the dataset.
Questions with tags \textit{[refer]} and \textit{[identify]} make up the majority of questions, approximately $67\%$ of the dataset.
\Cref{supp_fig:distribution} \textbf{(c)} further demonstrates the distribution across the parent datasets---EPIC-Kitchens and Ego4D.
Due to the availability of more videos and annotations in Ego4D, there are more QA pairs from Ego4D in \dataset over EPIC-Kitchens.
However, we do not have \textit{[vqa]} questions from Ego4D as it does not contain hand-object interaction information as provided in EPIC-Kitchens VISOR~\cite{VISOR2022}.
This also leads to less questions from Ego4D for \textit{[detect]}, \textit{[caption]}, and \textit{[grounding]}.

\Cref{supp_fig:distribution} \textbf{(d)} shows the distribution of the train and validation splits of EPIC-Kitchens and Ego4D across the proposed \dataset dataset.
As can be seen, the train set of \dataset contains a similar set of QA pairs from EPIC-Kitchens and Ego4D for training, however, due to a large number of videos in the validation set of Ego4D's FHO benchmark, the number of QA pairs from Ego4D in the test set of \dataset are greater than EPIC-Kitchens.

\subsubsection{Bounding Box Distribution Heatmaps.} 
\begin{figure}[t]
    \centering
    \includegraphics[width=\textwidth]{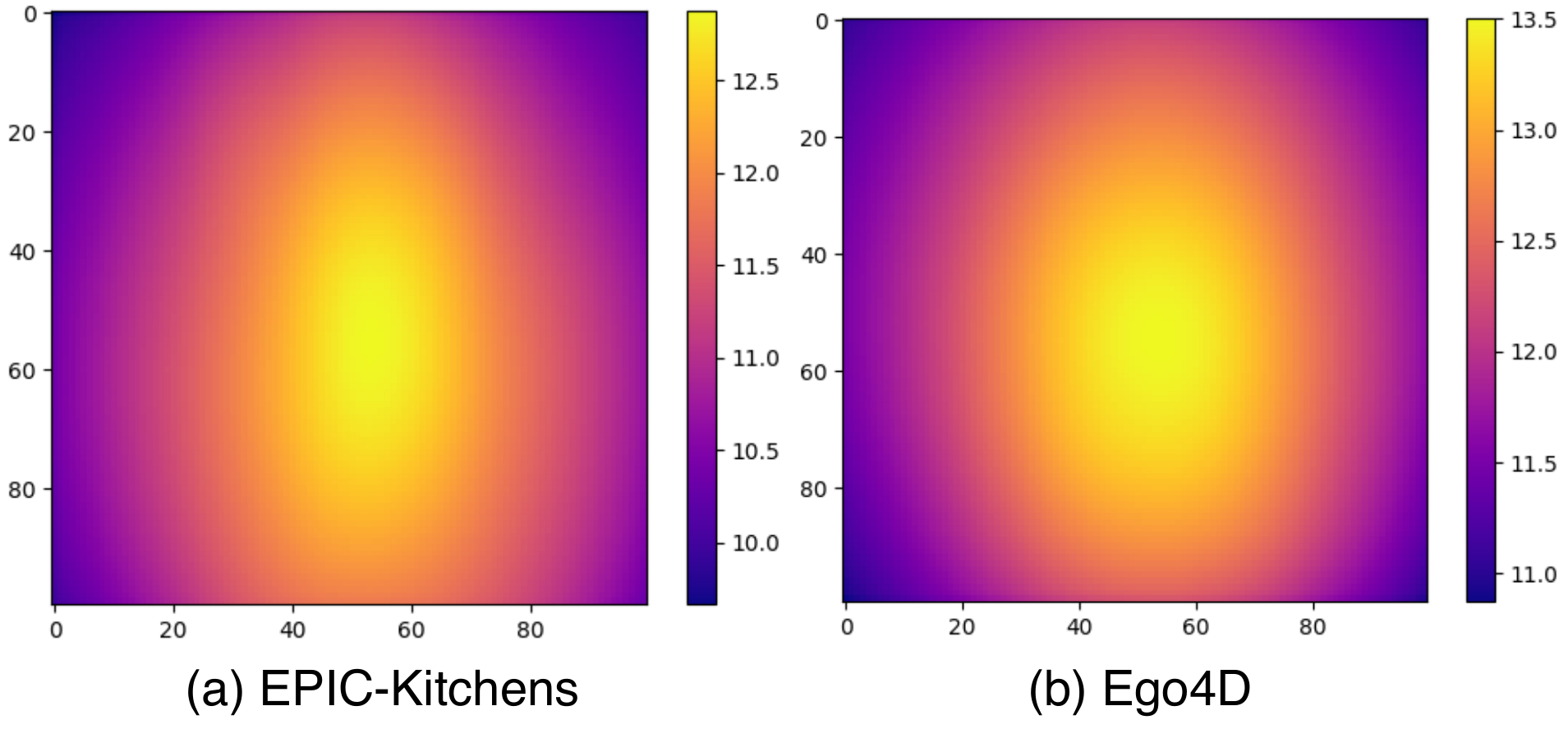}
    \caption{\textbf{Bounding Box's heatmaps:} \textbf{(a)} shows the heat map of bounding boxes in answers from EPIC-Kitchens and \textbf{(b)} shows the same for Ego4D used to generate the question-answer pairs for \dataset.}
    \label{supp_fig:heatmaps}
\end{figure}

To analyse the distribution of bounding boxes across the proposed \dataset dataset, we create heatmaps of the bounding boxes in EPIC-Kitchens and Ego4D.
To generate the heatmaps, we overlay all the bounding boxes on an $100\times100$ image and take the $\log$ of it.
\Cref{supp_fig:heatmaps} shows the two heatmaps.
Due to the nature of egocentric videos, the bounding boxes are close to the centre.
However, as we are considering bounding boxes for both hands and objects, the distribution is more spread.
Moreover, owing to the diversity of activities in Ego4D, the distribution of bounding boxes show a greater deviation in comparison to EPIC-Kitchens.
Finally, it is important to note that the \textit{Random} baseline from the main paper highlights that using the mean bounding box position/size (calculated from the training data) will lead to only $1.22\%$ bounding box accuracy.
Therefore, we can conclude that the bounding box distribution is not strongly biased towards the centre and well-performing models must understand the visual and textual aspects of the task to succeed.

\subsubsection{Bounding Box Size Analysis.}
\begin{figure}
    \centering
    \includegraphics[width=\textwidth]{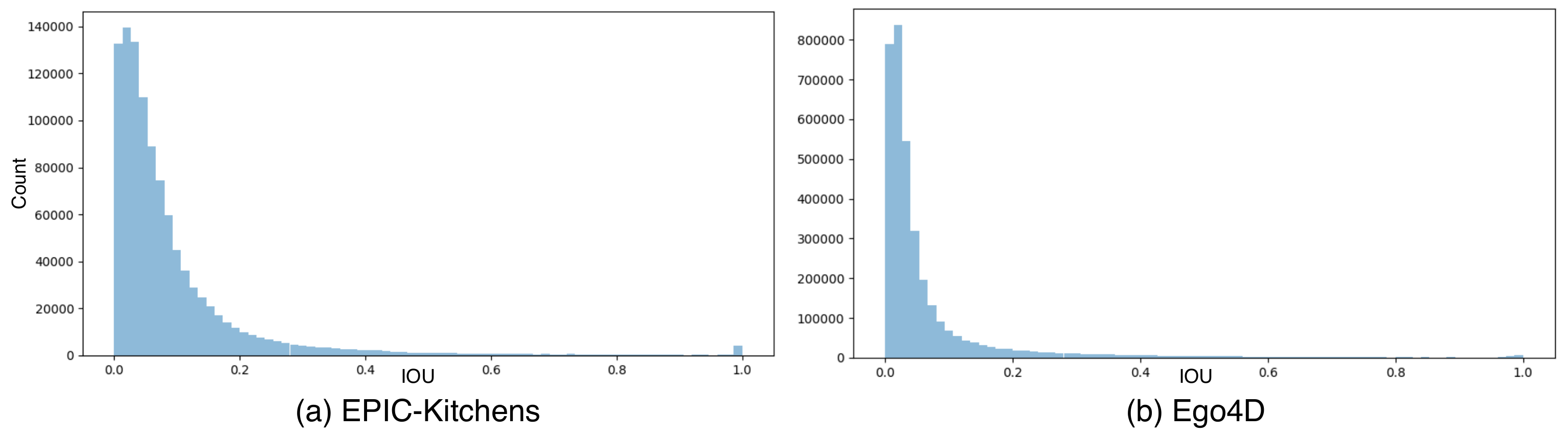}
    \caption{\textbf{Bounding Box Size Distribution.} The right skewed distribution shows that majority of the objects in the dataset are small in size.}
    \label{supp_fig:bbox_size}
\end{figure}

To understand the hand and object bounding box's size distribution in \dataset, we calculate the IOU between the image and the bounding box.
\Cref{supp_fig:bbox_size} shows the histogram of IOU distribution for data from EPIC-Kitchens and Ego4D in \dataset.
Both the histograms are right skewed showing that majority of the bounding boxes are small in the dataset.
Therefore, for a model to achieve high accuracy for referral, understanding small objects is crucial.

\subsubsection{Word Clouds for Nouns.} 
\begin{figure}[t]
    \centering
    \includegraphics[width=0.8\textwidth]{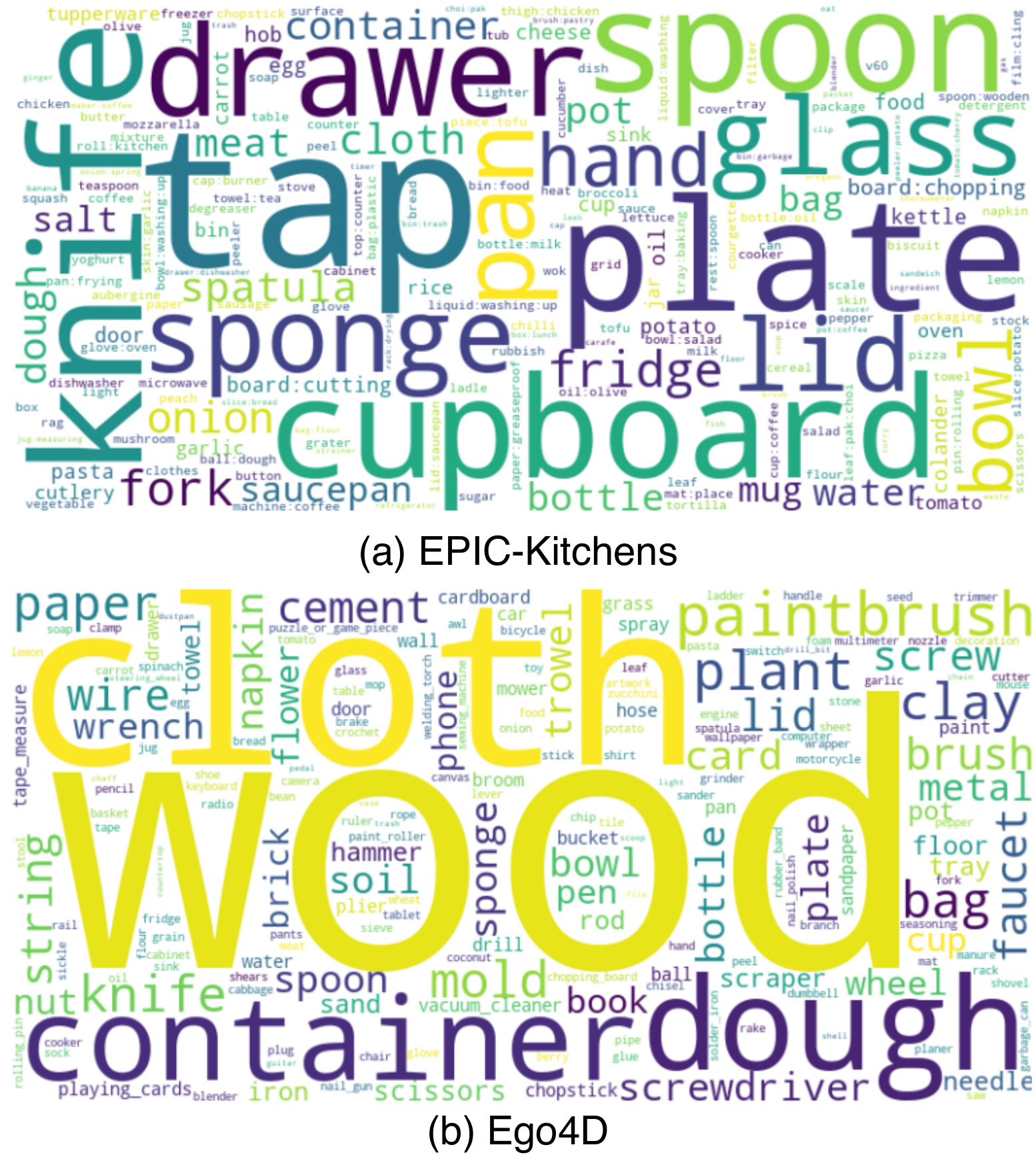}
    \caption{\textbf{\dataset Word cloud for nouns in EPIC-Kitchens and Ego4D.} Owing to the nature of datasets, we see more diverse vocabulary in Ego4D and a specific vocabulary in EPIC-Kitchens. Due to this, the overall vocabulary of the \dataset dataset is diverse.}
    \label{supp_fig:wordcloud}
\end{figure}

In \cref{supp_fig:wordcloud}, we show the word clouds of nouns in the proposed \dataset dataset from EPIC-Kitchens and Ego4D.
As can be seen in the figure, there is a clear difference between the two datasets due to the nature of their collection.
EPIC-Kitchens consists of nouns surrounding kitchen objects whereas Ego4D contains more diverse nouns.
Due to this, identifying objects and generating the nouns on Ego4D is challenging as also highlighted in \cref{supp_tab:per_dataset}.

\subsubsection{Images and Videos Count.} In \cref{supp_tab:img_count}, we show the count of images and videos used from EPIC-Kitchens and Ego4D to generate the QA pairs for the \dataset dataset.

\begin{table}[t]
    \centering
    \caption{
    \textbf{Number of images and videos} used from EPIC-Kitchens and Ego4D to generate $3.9$M question-answer pairs in the \dataset dataset
    }
    \setlength{\tabcolsep}{5pt}
    \begin{adjustbox}{width=0.78\textwidth}
    \begin{tabular}{@{}lcc@{}}\toprule
    Dataset & Number of Images & Number of Videos \\
    \midrule
    EPIC-Kitchens~\cite{Damen2022RESCALING,Damen2018EPICKITCHENS,VISOR2022}& $208,656$ & $495$ \\
    Ego4D~\cite{grauman2022ego4d}& $665,217$ & $2,461$ \\
    \bottomrule
    \end{tabular}
    \end{adjustbox}
    \label{supp_tab:img_count}
\end{table}

\subsection{List of all the question pairs to generate the data}
Apart from the question-answer pairs shown in the main paper, there are a variety of ways in which the same question has been phrased.
Here are all the prompts used as questions in the \dataset dataset divided based on the task tags:

\begin{enumerate}
    \item \textit{captions}
    \begin{enumerate}
        \item This is a photo from an ongoing video of hand object interactions. What is happening in the photo?
        \item What is going on in the photo?
        \item This is a photo has hand object interactions. Describe the ongoing action in the photo.
        \item What is the hand-object interaction happening in the image?
        \item What actions are the hands doing in this photo?
    \end{enumerate}
    \item \textit{vqa}
    \begin{enumerate}
        \item This is a photo from an ongoing video of hand object interactions. Is there an object in the left hand?
        \item Is the left hand holding or touching an object?
        \item This is a photo from an ongoing video of hand object interactions. Is the left hand holding an object or is object-free?
        \item Is the left hand object-free?
        \item This is a photo from an ongoing video of hand object interactions. Is there an object in the right hand?
        \item Is the right hand holding or touching an object?
        \item This is a photo from an ongoing video of hand object interactions. Is the right hand holding an object or is object-free?
        \item Is the right hand object-free?
        \item This is a photo from an ongoing video of hand object interactions. What is the object in the left hand?
        \item What is the object in the hand of the person?
        \item What is the person holding in their left hand?
        \item What is the object in the right hand?
        \item This is a photo from an ongoing video of hand object interactions. What is the person holding in their right hand?
        \item What is the object in the hands of the person?
        \item What hand is holding the \textit{\{object's name\}}?
        \item What hand is holding the \textit{\{object's name\}}?
        \item What hand is holding the \textit{\{object's name\}} in the image?
    \end{enumerate}
    \item \textit{identify}
    \begin{enumerate}
        \item \textit{${\{<X_{\text{left}}><Y_{\text{top}}><X_{\text{right}}><Y_{\text{bottom}}>\}}$}
    \end{enumerate}
    \item \textit{refer}
    \begin{enumerate}
        \item Where is the left hand of the person?
        \item Where is the left hand?
        \item Where is the left hand of the person in the image?
        \item left hand
        \item Where is the right hand of the person?
        \item Where is the right hand?
        \item Where is the right hand of the person in the image?
        \item right hand
        \item Where are the hands of the person?
        \item Where are the hands?
        \item Where are the hands of the person in the image?
        \item Locate the object being manipulated by the hand
        \item Where is the object being manipulated by the hand?
        \item Locate the manipulated object
        \item Where is the manipulated object?
        \item Which hand is holding the \textit{\{object's name\}} in?
        \item Which hand is holding the \textit{\{object's name\}} in the image?
        \item Which hand has the \textit{\{object's name\}}?
    \end{enumerate}
    \item \textit{grounding}
    \begin{enumerate}
        \item This is a photo from an ongoing video of hand object interactions. Describe hand-object interaction in detail
        \item Describe the ongoing action in this image in detail
        \item This is a photo from an ongoing video of hand object interactions. Describe what actions are the hand doing in this photo
        \item This is a photo from an ongoing video of hand object interactions. Describe the ongoing action in this image
        \item Describe the ongoing action in this image
    \end{enumerate}
    \item \textit{detection}
    \begin{enumerate}
        \item \textit{\{object's name\}}
    \end{enumerate}
\end{enumerate}

\section{Additional Quantitative Results}
\label{supp_sec:quantiative_results}

\subsection{Analysis on \hobenchmark and \hoibenchmark}

Here we take a deeper look into the results for \benchmark on \dataset from \hobenchmark and \hoibenchmark's perspective.
We do this by further dividing the results into the four categories of question-answer types described in Section 3.3 in the main paper.
The results are divided across two tables, \cref{supp_tab:ho_categories} and \cref{supp_tab:i_categories}.

\begin{table}[t]
    \centering
    \caption{
    \textbf{Results on subsets in \hobenchmark} showcases the ability of baseline and \model on each category of QA pairs.
    Here, Obj N stands for Object Noun, Obj BB stands for object bounding box, Hand S stands for Hand Side (Left/Right), and Hand BB stands for hand bounding box
    }
    \setlength{\tabcolsep}{5pt}
    \begin{adjustbox}{width=1\textwidth}
    \begin{tabular}{@{}lcccc@{}}\toprule
    Model & Obj N $\rightarrow$ Obj BB & Hand S $\rightarrow$ Hand BB & Obj BB $\rightarrow$ Obj N & Hand BB $\rightarrow$ Hand S \\
    \midrule
    MiniGPT-v2~\cite{chen2023minigptv2}& $25.37$ & $19.90$& $15.67$& $0.01$ \\
    \model(ours)& $\mathbf{34.27}$ & $\mathbf{60.38}$ & $\mathbf{31.85}$& $\mathbf{5.60}$ \\
    \bottomrule
    \end{tabular}
    \end{adjustbox}
    \label{supp_tab:ho_categories}
\end{table}

As seen in \cref{supp_tab:ho_categories}, \model outperforms baseline in all the four categories.
\model has high performance when localising the object or hands showing its superior capability for referral in egocentric images.
Moreover, the baseline falls short on generating hand side's name provided hand's bounding box, whereas, \model has $5.59\%$ improvement over it.

\Cref{supp_tab:i_categories} shows similar results for \hoibenchmark.
Here, we test the VLMs capability to refer to object/hand provided hand/object information.
As can be seen, \model significantly outperforms the baseline.
Especially for the case where object name is provided and hand's bounding box or noun is supposed to be generated, \model performs $43.41\%$ better on average.
Furthermore, when object's name is provided, generating hand's name is challenging for MiniGPT-v2 however, \model performs well on such cases.

\begin{table}[t]
    \centering
    \caption{
    \textbf{Results on subsets in \hoibenchmark} showcases the ability of baseline and \model on each category of QA pairs.
    Here, Obj N stands for Object Noun, Obj BB stands for object bounding box, Hand S stands for Hand Side, and Hand BB stands for hand bounding box
    }
    \setlength{\tabcolsep}{5pt}
    \begin{adjustbox}{width=1\textwidth}
    \begin{tabular}{@{}lcccc@{}}\toprule
    Model & Hand S $\rightarrow$ Obj BB & Obj N $\rightarrow$ Hand BB & Hand S $\rightarrow$ Obj N & Obj N $\rightarrow$ Hand S\\
    \midrule
    MiniGPT-v2~\cite{chen2023minigptv2}& $27.69$& $32.55$& $14.12$& $0.00$ \\
    \model(ours)& $\mathbf{41.13}$& $\mathbf{72.59}$& $\mathbf{28.11}$& $\mathbf{46.78}$ \\
    \bottomrule
    \end{tabular}
    \end{adjustbox}
    \label{supp_tab:i_categories}
\end{table}

\subsection{Results per parent dataset: EPIC-Kitchens and Ego4D}

In \cref{supp_tab:per_dataset} we show the results the validation splits of EPIC-Kitchens and Ego4D used to create the test set of \dataset.
There is a large boost in performance when compared to MiniGPT-v2~\cite{chen2023minigptv2} on both EPIC and Ego4D.
The boost in noun accuracy for Ego4D is less compared to EPIC-Kitchens due to the diversity of nouns in Ego4D as shown in \cref{supp_fig:wordcloud}.

\begin{table}[t]
    \centering
    \caption{
    \textbf{Results on EPIC-Kitchens and Ego4D in \dataset.}
    Our \model outperforms MiniGPT-v2 on both the subsets within \dataset
    }
    \setlength{\tabcolsep}{5pt}
    \begin{adjustbox}{width=1\textwidth}
    \begin{tabular}{@{}lcccccc@{}}\toprule
    \multirow{2}{*}{Model} & \multicolumn{3}{c}{EPIC-Kitchens~\cite{Damen2018EPICKITCHENS,Damen2022RESCALING,VISOR2022}} & \multicolumn{3}{c}{Ego4D~\cite{grauman2022ego4d}}\\
    \cmidrule(lr){2-4}\cmidrule(l){5-7}
    & Noun Acc. & Bounding Box Acc. & Avg. BB IOU & Noun Acc. & Bounding Box Acc. & Avg. BB IOU \\
    \midrule
    MiniGPT-v2~\cite{chen2023minigptv2}& $13.56$ & $34.88$& $33.96$& $5.38$& $19.56$& $22.04$ \\
    \model(ours)& $\mathbf{43.32}$ & $\mathbf{65.30}$& $\mathbf{58.24}$& $\mathbf{12.86}$& $\mathbf{46.76}$& $\mathbf{39.76}$ \\
    \bottomrule
    \end{tabular}
    \end{adjustbox}
    \label{supp_tab:per_dataset}
\end{table}

\section{Zero-shot Interactive Results}
\label{supp_sec:qualitative_results}

In this section, we perform zero-shot qualitative analysis using the \model trained on \dataset.
For this purpose, we randomly select images from the test set of \dataset and prompt the model with novel questions interactively.
As there are no ground truth for these questions, we qualitatively assess the generation.
This highlights the generalisability of \model over various prompts for hand-object interaction referral on egocentric images.

\begin{figure}[!tbhp]
    \centering
    \includegraphics[width=0.93\textwidth]{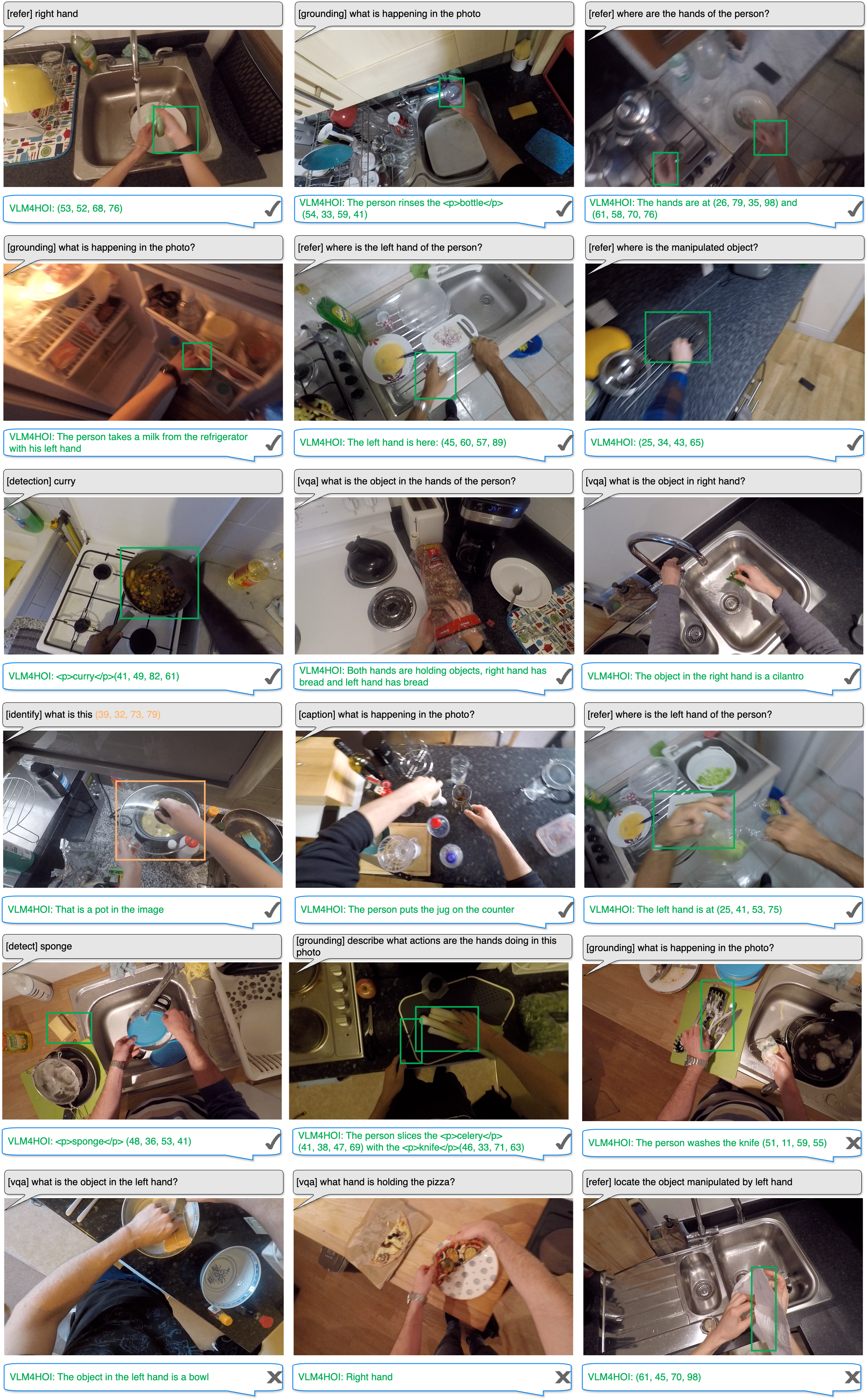}
    \caption{Zero-shot results on the EPIC-Kitchen's subset of \dataset validation set using \model. Green bounding boxes represent \model's output and orange bounding boxes represent input to the model.
    After qualitative analysis the correct generations are marked by a tick and wrong generations are marked by a cross.}
    \label{supp_fig:epic_qual}
\end{figure}

\begin{figure}[!tbhp]
    \centering
    \includegraphics[width=0.7\textwidth]{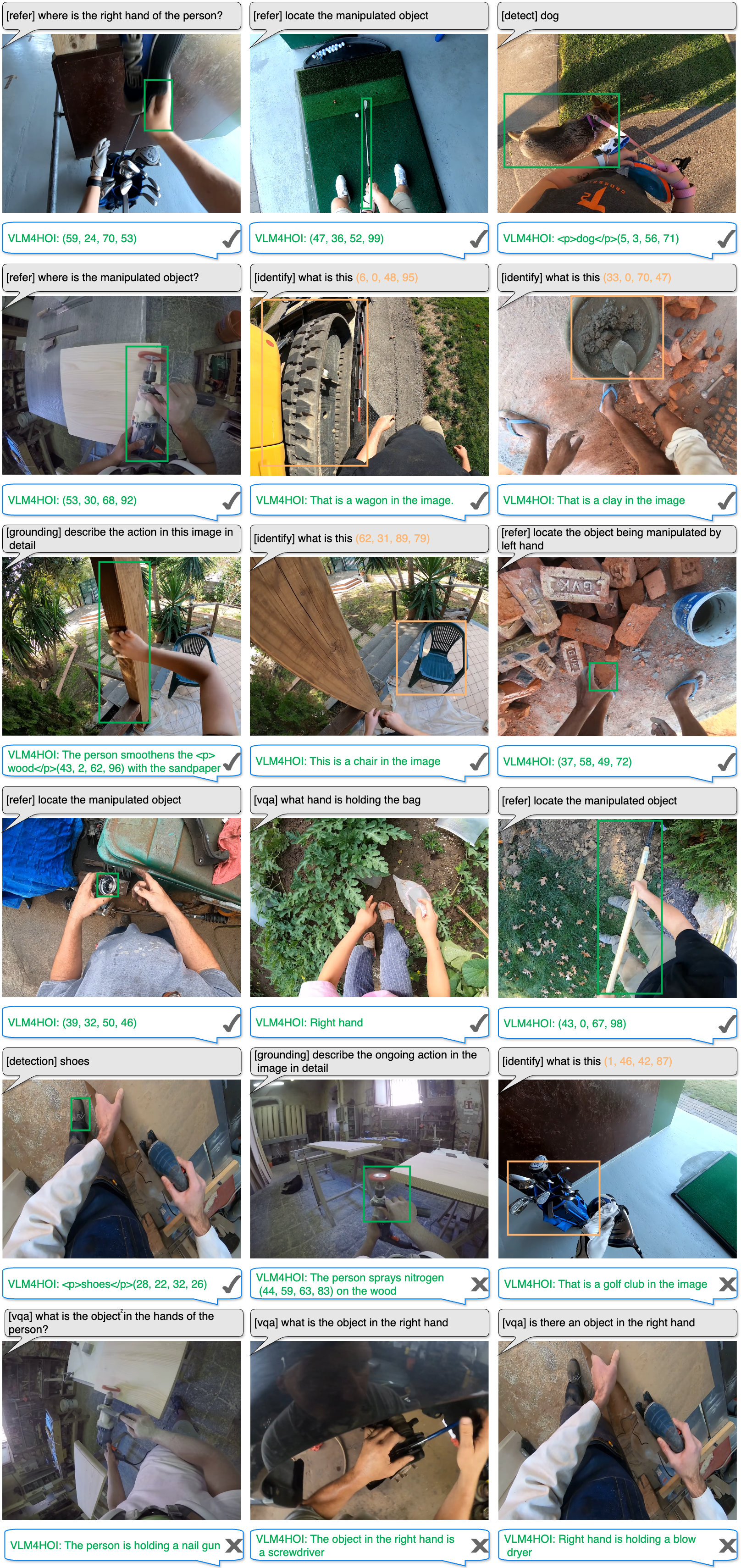}
    \caption{Zero-shot results on the Ego4D's subset of \dataset validation set using \model.
    Green bounding boxes represent \model's output and orange bounding boxes represent input to the model.
    After qualitative analysis the correct generations are marked by a tick and wrong generations are marked by a cross.}
    \label{supp_fig:ego4d_qual}
\end{figure}

\Cref{supp_fig:epic_qual} shows the results on images from the EPIC-Kitchens used in \dataset.
As can be seen in the figure, the model generates correct results for majority of the cases.
For example, the model correctly identifies the hands, objects in hands, and generates the name of the objects.
It confuses for the cases where the objects are overlapping, for example, in row $6$ column $1$, the model confuses between the fish and bowl.
\Cref{supp_fig:ego4d_qual} presents the results on the images from the Ego4D used in \dataset.
Here, the model is able to refer to various objects in the wild for example, golf club, dog, and a wagon.
However, it confuses for the cases where objects are similar looking for example, in row $5$ column $2$, the model confuses sanding the wood with spraying nitrogen on the wood.

\end{document}